\documentclass[letterpaper, 10 pt, conference]{ieeeconf}  

\IEEEoverridecommandlockouts                              

\usepackage{graphics} 
\usepackage{epsfig} 
\usepackage{mathptmx} 
\usepackage{times} 
\DeclareMathAlphabet{\mathcal}{OMS}{cmsy}{m}{n}
\usepackage{subfig}
\usepackage{algorithm}
\usepackage{algpseudocode}
\usepackage{amsmath} 
\usepackage{amssymb}  
\usepackage{hyperref}
\hypersetup{
colorlinks=true,
linkcolor=black,
hyperindex=true,
citecolor=red}
\usepackage{cite}
\newcommand{\algname}{\mbox{TPP}}

\usepackage{color}
\usepackage{pifont}
\usepackage{multirow}

\DeclareMathOperator{\bfr}{\mathbf{r}}
\DeclareMathOperator{\bfe}{\mathbf{e}}

\DeclareMathOperator{\pa}{\text{Par}}
\DeclareMathOperator{\ch}{\text{Ch}}

\DeclareMathOperator{\ck}{{\color{green}\checkmark}}
\DeclareMathOperator{\xm}{{\color{red}\mathbf{\times}}}

\newtheorem{definition}{Definition}
\newtheorem{remark}{Remark}

\newtheorem{thm}{Theorem}

\newtheorem{prop}{Proposition}

\usepackage{xcolor}
\newcommand{\comment}[1]{{\color{blue}Comment: #1}} 
\newcommand{\TODO}[1]{{\color{red}TODO: #1}} 
\renewcommand{\comment}[1]{} 
\renewcommand{\TODO}[1]{} 

\newcommand{\myparagraph}[1]{\noindent\textbf{#1}~} 

\title{\LARGE \bf
Tree-structured Policy Planning with Learned Behavior Models
}

\author{Yuxiao Chen$^{1}$, Peter Karkus$^{1}$, Boris Ivanovic$^{1}$, Xinshuo Weng$^{1}$, and Marco Pavone$^{1,2}$
\thanks{$^{1}$The authors are with NVIDIA Research, NVIDIA Corporation, 2788 San Tomas Expressway, Santa Clara, CA.
        {\tt\small \{yuxiaoc,pkarkus,bivanovic,xweng\}@nvidia.com}}
\thanks{$^{2}$Marco Pavone is with the Department of Aeronautics and Astronautics, Stanford University, and with NVIDIA Research.
        {\tt\small \{pavone@stanford.edu, mpavone@nvidia.com\}}}%
}

\begin{document}

\maketitle
\thispagestyle{empty}
\pagestyle{empty}

\begin{abstract}
Autonomous vehicles (AVs) need to reason about the multimodal behavior of neighboring agents while planning their own motion. Many existing trajectory planners seek a single trajectory that performs well under \emph{all} plausible futures simultaneously, ignoring bi-directional interactions and thus leading to overly conservative plans. Policy planning, whereby the ego agent plans a policy that reacts to the environment's multimodal behavior, is a promising direction as it can account for the action-reaction interactions between the AV and the environment. However, most existing policy planners do not scale to the complexity of real autonomous vehicle applications: they are either not compatible with modern deep learning prediction models, not interpretable, or not able to generate high quality trajectories. To fill this gap, we propose Tree Policy Planning (\algname{}), a policy planner that is compatible with state-of-the-art deep learning prediction models, generates multistage motion plans, and accounts for the influence of ego agent on the environment behavior. The key idea of \algname{} is to reduce the continuous optimization problem into a tractable discrete Markov Decision Process (MDP) through the construction of two tree structures: an ego trajectory tree for ego trajectory options, and a scenario tree for multi-modal ego-conditioned environment predictions.  We demonstrate the efficacy of \algname{} in closed-loop simulations based on real-world nuScenes dataset and results show that \algname{} scales to realistic AV scenarios and significantly outperforms non-policy baselines.

\end{abstract}

\section{Introduction}\label{sec:intro}

A key challenge of motion planning for autonomous vehicles (AV)s is reasoning about the interaction between the ego vehicle and  neighboring agents. 
The task is commonly divided into two subproblems: trajectory prediction for other agents, and ego motion planning with prediction. 
Trajectory prediction has seen substantial progress in recent years, coming from simple kineamtic models \cite{kong2015kinematic} to powerful deep learning models \cite{predictionnet,T++,rhinehart2019precog,ILVM,ScePT} capable of generating high-quality, multi-modal predictions. However, motion planning with such high-capacity prediction models remains a challenge.

Typical AV planners simplify the ego behavior planning problem into \emph{trajectory planning}, where a single trajectory is sought that minimizes an expected cost over \emph{all} predicted futures within the planning horizon~\cite{xu2014motion}. 
Such trajectory planning leads to overly conservative plans because it ignores the new information to be acquired about other agents in subsequent time steps, and the effects of ego actions on the behavior of other agents.

In fact, in decision-making literature, people have long been pursuing a closed-loop policy instead of an open-loop plan as the former is shown to be optimal for problems such as Markov decision processes (MDP) \cite{puterman1990markov}. 

\bgroup
\setlength{\tabcolsep}{4pt}
\begin{table}[b]
\vspace{-0.4cm}
\begin{tabular}{l|ccccc}& DL & EC  & Interp & MS & Scalability \\\hline
Multi-mode planning
\cite{hardy2010contingency, hardy2013contingency,zhan2016non,galceran2017multipolicy} &  $\xm$  &  $\xm$  & $\ck$ &$\xm$ & poor \\
Branch MPC \cite{chen2022interactive} & $\xm$  & $\ck$ & $\ck$&  $\ck$ & poor\\
POMDP \cite{bai2015intention,liu2015situation,intentionaware}  & $\xm$ &  $\ck$  &$\ck$ &$\ck$ & poor\\

Game theoretic \cite{fisac2019hierarchical,sadigh2016planning}  & $\xm$   &   $\ck$  & $\ck$ &$\xm$  & very poor     \\
Lookout \cite{cui2021lookout}    &   $\ck$ &  $\xm$   &      $\ck$  &$\xm$    & good   \\
CfO \cite{CfO}   &$\ck$  & $\ck$ &  $\xm$ & $\xm$ & medium \\
\algname{} (ours)  & $\ck$ &  $\ck$  &   $\ck$ & $\ck$ & good
\end{tabular}
\caption{Comparison of existing policy planning methods and \algname{} where ``DL'' is short for compatibility with deep-learned prediction models, ``EC'' stands for ego-conditioning, ``Interp'' is short for interpretable, ``MS'' stands for multistage, and ``Scalability'' means the capability to scale to complicated scenes with large number of agents.}
\label{table:exist_method}
\vspace{-0.5cm}
\end{table}

\egroup

\begin{figure*}
    \centering
    \includegraphics[width=0.9\textwidth]{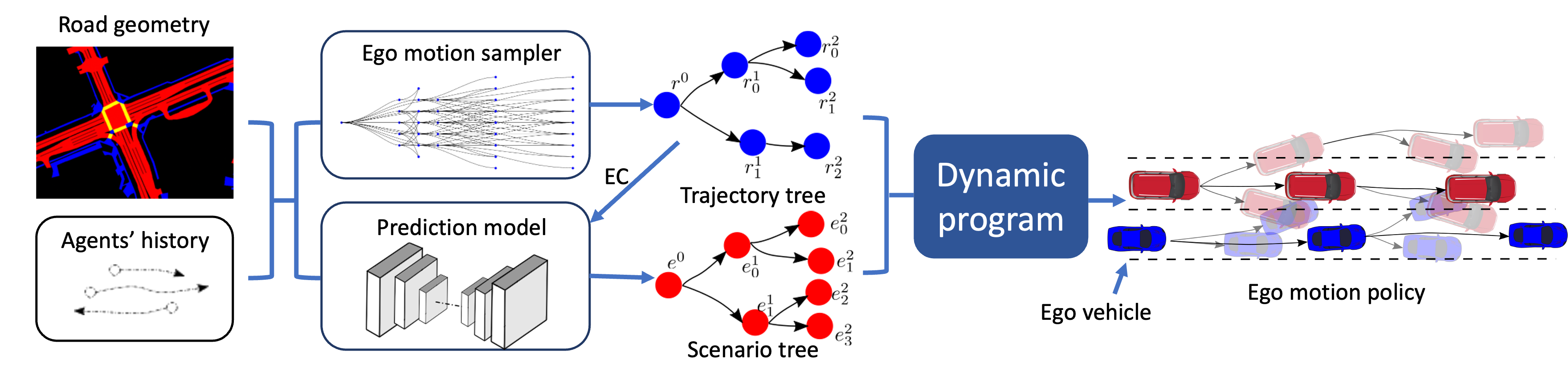}
    \caption{Structure of \algname{}: the ego motion sampler generates the trajectory tree, which is fed to the prediction model for ego-conditioned prediction to generate the scenario tree. Then we solve for the ego motion policy via dynamic programming.}
    \label{fig:diagram}
    \vspace{-0.6cm}
\end{figure*}

To highlight the difference between a \emph{motion policy} and a single trajectory, consider the scenario depicted on the right end of Fig. \ref{fig:diagram}. The ego vehicle (blue) needs to move pass the adjacent vehicle (red). Within the planning horizon, the red vehicle either maintains its current lane or cuts in front of the ego vehicle. A traditional planner seeks a single trajectory that performs well in both situations and thus will not choose to pass since lane change is a plausible choice for the red vehicle at current time. Alternatively, a motion policy may choose to nudge forward for the first part of the horizon, observe (by leveraging a forecasting model) the behavior of the adjacent vehicle, and then decide the subsequent motion. If the red vehicle already started a lane change, the ego vehicle brakes to avoid collision; if the red vehicle stays in its own lane, since the ego vehicle is already next to the red vehicle, a sudden lane change is very unlikely, the ego vehicle can then move past the red vehicle. 

In the field of robotics, policy planning often takes the form of \emph{contingency planning}~\cite{hardy2013contingency,zhan2016non}, with the key idea of preparing multiple plans for different likely futures. To differentiate the two concepts, we view contingency planning as a simplified policy planning variant with only one stage of reasoning. Unfortunately, continuous-space policy planning with high-capacity prediction models is computationally prohibitive, and despite various simplifications ~\cite{chen2022interactive,cui2021lookout,bai2015intention}, practical policy planning for AVs remains an open problem.

In this paper we introduce Tree Policy Planning (\algname{}), a practical policy planner that meets the requirements of real-world autonomous driving, specifically, \algname{} is designed with the following desiderata:
\begin{enumerate}
    \item compatible with state-of-the-art deep-learned predictive models for trajectory forecasting 
    \item plans with closed-loop ego-conditioning, i.e., factors in the ego vehicle's influence on other agents
    \item allows for multistage reasoning that leverages the agents' future reactive behavior
    \item interpretable and easily tunable 
    \item scales to scenarios with a large number of agents
\end{enumerate}
We compare existing policy planning works with TPP in Table \ref{table:exist_method}. To the best of our knowledge, TPP is the first algorithm that fulfils our desiderata.

The key idea of \algname{} is to convert the continuous space motion planning problem into a tractable finite-horizon MDP by constructing two trees: an \emph{ego trajectory tree} carrying ego motion candidates, and a \emph{scenario tree} that contains multiple behavior modes for neighboring agents. The algorithm is illustrated in Fig.~\ref{fig:diagram}. It consists of three components: an ego motion sampler that generates the ego trajectory tree, a deep-learned multi-stage prediction model that generates the scenario tree, and a dynamic program module that solves for the optimal policy with respect to the constructed MDP. 

We evaluate \algname{} in closed-loop simulation based on a recently proposed realistic data-driven traffic model~\cite{BITS}. Our results show that \algname{} achieves good closed-loop driving performance in real-world scenarios from the nuScenes dataset \cite{nuscenes}, and significantly outperforms non-policy planning baselines. The algorithm is highly parallelizable, and can run in real time.

\section{Related works}\label{sec:related_work}

Various contingency planning algorithms have been proposed and they can be viewed as different approximations of policy planning. A complete policy planning algorithm should be capable of multi-stage reasoning with bi-directional interactions between the agent and the environment. Since the ego is always reacting to the environment, the concept of bi-directional interaction is also referred to as ego-conditioning, emphasizing on the environment reacting to the ego agent. The application of autonomous driving also asks for high quality trajectories, accurate prediction, interpretability, and scalability. All these features combined makes it difficult to realize a non-compromising policy planning algorithm for autonomous vehicles. As a result, different simplification methods have been proposed. Limited to one stage, \cite{hardy2013contingency} proposed a trajectory optimization method with one trajectory for each scene evolution mode, similar ideas are also seen in \cite{zhan2016non,galceran2017multipolicy} where a predefined set of modes are evaluated online. \cite{cui2021lookout} uses a deep-learned prediction model to generate the modes. \cite{chen2022interactive} performs multistage reasoning via branch MPC and leveraged ego-conditioning, yet requires a simple differentiable prediction model. 

Partially Observed Markov Decision Process (POMDP) ~\cite{bai2015intention,liu2015situation,intentionaware} is a classic policy planning method, yet cannot generate high quality trajectories or use powerful deep-learned prediction models. Game theoretic approaches are well-known for reasoning about bi-directional interactions, yet assume some simple fixed behavior model of the environment, e.g. rational \cite{sadigh2016planning} or noisy Boltzmann \cite{fisac2019hierarchical}. \cite{rhinehart2021contingencies} performs policy planning by training a neural network to react to the environment behavior, yet gives up on multistage reasoning, and is not interpretable and hard to tune. 

Table \ref{table:exist_method} summarizes the various existing methods and their compromises, and we would like to show in the remainder of the paper that \algname{} closes the gap to a practical policy planning method for AV.

\section{Tree Policy Planning}\label{sec:formulation}

\algname{} generates the ego motion policy in three steps as shown in Fig.~\ref{fig:diagram}. First, the ego motion sampler generates an \emph{ego trajectory tree}, $\bfr$, given the current ego state and a lane map (when available). The ego trajectory tree consists of connected, dynamically feasible trajectory segments that together represent candidate ego-vehicle trajectories the planner can choose from. Second, the prediction module is a deep learned model that generates multi-modal predictions of the environment in the form of a \emph{scenario tree}, $\bfe$. 
Finally, the dynamic programming module 
converts $\bfr$ and $\bfe$ into an MDP and finds an optimal policy via dynamic programming. 

We use the following notation. Let $s_r$ denote the state of the ego vehicle and $s_e$ denote the state of the environment, including the state of nearby agents and the location of static obstacles. 
For the trajectory tree, $r^i_j$ denotes the $j$-th node of stage $i$ and and the root node is simply $r^0$. The nodes in the scenario tree are labeled the same way. We enforce that all nodes of the same stage contain trajectories of the same duration with $t^i_0$ and $t_f^i$ being the starting and ending time of the $i$-th stage. The trajectory of a any node other than the root needs to follow the trajectory of its parent node. For both trees, $\pa(\cdot)$ denotes the parent of a node and $\ch(\cdot)$ denote the set of children of a node, e.g., $\pa(r_1^1)=r^0$, $\ch(e_0^1)=\{e^2_0,e^2_1\}$ for the two trees in Fig. \ref{fig:diagram}. 


Next, we discuss the three components in detail.

\subsection{Ego-motion sampler}\label{sec:sampler}
The ego motion sampler generates a set of dynamically feasible trajectory options for planning to consider, and organizes them into a trajectory tree. 

\noindent\myparagraph{Terminal point sampling.} We first sample a set of terminal points and then find a feasible trajectory to the terminal point.
Terminal points are sampled from a fixed set of acceleration-steering values, as well as along (and near to) lane centerlines when lane information is available. 

\noindent\myparagraph{Spline fitting.} We then fit a spline from the current ego state to the terminal point following \cite{schmerling2018multimodal}.
\comment{the rest can be removed if needed}
Given the current ego state $s_r[0]=[X_0,Y_0,v_0,\psi_0]^\intercal$ and a terminal condition $s_r[T]=[X_T,Y_T,v_T,\psi_T]^\intercal$, we use two separate polynomials $X[t]$ and $Y[t]$ to parameterize the $X,Y$ coordinates such that the resulting spline connects the two points and the boundary conditions are met:
\begin{equation}
\begin{aligned}
    X[0]&=X_0\quad &X[T]&=X_T\\
    Y[0]&=Y_0\quad &Y[T]&=Y_T\\
    \dot{X}[0]&=v_0 \cos\psi_0\quad &\dot{X}[T]&=v_T \cos\psi_T\\
    \dot{X}[0]&=v_0 \sin\psi_0\quad &\dot{X}[T]&=v_T \sin\psi_T.
\end{aligned}
\end{equation}
With four boundary conditions on each polynomial, two 3rd order (cubic) polynomial can be uniquely determined, which allows us to directly compute the coefficient of $X[t]$ and $Y[t]$ in closed-form and calculate the trajectory from $X[t]$ and $Y[t]$. 
Finally we drop trajectories that violate dynamic constraints. 

\noindent\myparagraph{Tree construction.} In practice, trajectory trees are grown stage by stage, starting from the current state as root $r^0$. We set an upper bound on the number of children a node can have, and randomly drop children nodes if the limit is exceeded. 
In each stage we grow children branches in parallel, thus the time complexity is linear in stage number.

\subsection{Multi-stage prediction model}\label{sec:pred}
A critical part of \algname{} is the prediction model that generates a scenario tree. For policy planning, we need a scene-centric, ego-conditioned, and multi-stage prediction model. We will first review these requirements and then show examples of deep-learning models satisfying our requirements.

\noindent\myparagraph{Requirements.} 
First, the prediction model needs to predict \emph{multiple modes} of the trajectory distribution for each agent. Second,  predictions must be \textit{scene-centric}, i.e., outputting modes of the joint trajectory distribution of all relevant agents in the scene. Branches of our scenario tree corresponds to modes of the joint distribution so that our planner will be able to evaluate ego trajectory candidates against joint futures effectively. Examples for multi-modal scene-scentric predictions include~\cite{agentformer,ScePT}.
Third, we need \textit{ego-conditioning} (EC), i.e., prediction results should be conditioned on the ego vehicle's hypothetical future motion. Ego conditioning allows the planner to leverage the reactive behaviors of the environment towards the ego vehicle and is a known technique in prediction literature \cite{T++,ScePT}. However, training a EC prediction model remains challenging as the ground truth under counterfactual ego future motion is not known.

Lastly, we need to generate modes of the predicted trajectory distribution in \emph{multiple stages}. That is, the generated scenario tree contains multiple stages with each node branching into multiple children after each stage. Multistage prediction allows the ego agent to depend its future motion on the future multimodal observation of the environment (similar to a decision tree) and the stage number can be freely set as a design choice. Details on converting existing models to enable multistage prediction can be found in the appendix.

\noindent\myparagraph{Model architectures.} To evaluate \algname{}'s flexibility w.r.t. different prediction models, we tested it with three vastly distinct prediction models. The first model is a rasterized model with CNN backbone and a CVAE; the second model is the PredictionNet \cite{predictionnet} model with a Unet backbone, and the third model is the Agentformer model \cite{agentformer}. All models are modified to enable multistage prediction and ego-conditioning. Details can be found in the appendix.

\subsection{Dynamic programming}\label{sec:DP}
The dynamic programming module takes the ego trajectory tree $\bfr$ and the scenario tree $\bfe$, along with a handcrafted cost function $l$ and finds an optimal ego motion policy. 

Intuitively, our algorithm constructs a discrete MDP over $\bfr$ and $\bfe$ and finds an optimal policy through dynamic programming. More specifically, states of the MDP are nodes 
of the scenario/trajectory trees that represent the joint ego and environment state; transitions are given by the tree parent/children edges in the two trees; and rewards are defined by the integral of the cost over the time segment. We then solve for the optimal finite-horizon \emph{policy} for the discrete MDP using dynamic programming, i.e., calculating the value function backwards in stage via the Bellman equation. With the optimal policy, the continuous-time trajectory is recovered from the trajectory tree by executing the trajectory segments according to the policy. In practice we use a cost function $l$ with standard terms for collision avoidance, lane keeping, goal reaching, and ride comfort; and a dynamically extended unicycle model~\cite{lavalle2006better} for the vehicle dynamics.

Next we provide a formal discussion of the \algname{} algorithm.


\section{Formal discussion}\label{sec:formal}

\subsection{Analysis without ego-conditioning}
In our motion planning problem we are interested in minimizing the expected cumulative cost over a horizon:
\begin{equation*}
    \mathcal{J} = \int_{0}^T l(s(t),s_e(t))dt,
\end{equation*}
where $l$ is the running cost function that depends on the ego state $s$ and the environment state $s_e$, $T$ is the planning horizon.  Our planning algorithm operates on the two trees $\bfr$ and $\bfe$ with the same number of stages, so the cumulative cost can be rewritten as 
\begin{equation}\label{eq:stage_cumu_cost}
    \mathcal{J} = \sum_{i=0}^{N} L_i(r^i,e^i),
\end{equation}
where $r^i$ and $e^i$ are the nodes the ego vehicle and the environment take at stage $i$, and $L_i(r^i,e^i)=\int_{t^i_0}^{t^i_f} l(s^{r^i}(t),s_e^{e^i}(t)) dt$. Due to the stochasticity of the environment, we aim to minimize $\mathbb{E}_{e^{0:N}}[\sum_{i=0}^{N} L_i(r^i,e^i)]$, where the expectation is on $e^i$, the actual branch the environment takes and the distribution is given as part of the prediction. For simplicity, we shall present the result assuming $\bfe$ is not ego-conditioned, i.e., there is a single $\bfe$, and later show that the result can be easily extended to the case with ego-conditioning.

The end result is a policy that chooses which node to execute for the next stage given the current ego and environment node, i.e., $r^{i+1} = \pi^\star(r^i,e^i)$, where $\pi^\star$ is the optimal policy. The form of the policy is determined by the assumption about the information flow, i.e., during stage $i$, the ego vehicle can observe the environment's response $e^i$, and choose the ego motion among the children nodes of the current node $r^i$. To obtain the optimal policy, a value function is needed:
\begin{equation}\label{eq:V}
    V(r^i,e^i)=\mathbb{E}_{e^{i:N}} [\sum_{k=i}^N L_k(r^k,e^k)],
\end{equation}
which is the expected cost-to-go. The dynamic program goes backwards in stage and is trivial for the last stage:
\begin{equation*}
    V(r^N,e^N)=L_N(r^N,e^N).
\end{equation*}
For stage i where $i<N$, we have
\begin{equation*}
    V(r^i,e^i) = L_i(r^i,e^i)+\min_{r^{i+1}\in \ch(r^i)}\mathbb{E}_{e^{i+1}}[V(r^{i+1},e^{i+1})|e^i].
\end{equation*}
Since $e^{i+1}$ is enumerable, the expectation is calculated as 
\begin{equation*}
    \mathbb{E}_{e^{i+1}}[V(r^{i+1},e^{i+1})|e^i] = \sum_{e^{i+1}_j\in\ch(e^i)} \mathbb{P}[e^{i+1}_j|e^i] V(r^{i+1},e^{i+1}_j),
\end{equation*}
where the branching probability of the scenario tree is given by the prediction model. Further define
\begin{equation*}
\begin{aligned}
    Q(r^{i+1},e^i)&:=L_i(r^i,e^i)+\mathbb{E}_{e^{i+1}}[V(r^{i+1},e^{i+1})|e^i]\\ &= L_i(r^i,e^i)+\sum_{e^{i+1}_j\in\ch(e^i)} \mathbb{P}[e^{i+1}_j|e^i] V(r^{i+1},e^{i+1}_j).
    \end{aligned}
\end{equation*}
where $r^i=\pa(r^{i+1})$.
We name this function $Q$ function as it is analogous to the $Q$ function in reinforcement learning where here $b^{k+1}$ is the action we need to choose.

The value function is then obtained backwards in stage:
\begin{equation}
    V(r^i,e^i)=\left\{\begin{array}{cc}L_i(r^i,e^i),&i=N\\\min_{r^{i+1}\in \ch(r^i)}Q(r^{i+1},e^i),&i<N \end{array}\right.
\end{equation}
The optimal policy is then
\begin{equation}\label{eq:opt_policy}
    \pi^\star(r^i,e^i)=\mathop{\arg\min}_{r^{i+1}\in\ch(r^i)} Q(r^{i+1},e^i).
\end{equation}

\noindent\begin{thm}[Optimality]\label{thm:DP}
The policy in \eqref{eq:opt_policy} is the optimal policy with respect to the cost in \eqref{eq:stage_cumu_cost}.
\end{thm}
\noindent\begin{proof}
The proof is by induction. For stage $N$, the cost is a constant, $\pi^\star$ is trivially optimal, and $V(r^N,e^N)$ is the cost-to-go. For $0\le i<N$, assume $V(r^{i+1},e^{i+1})$ is the expected cost-to-go function at stage $i+1$, then by the principle of optimality, the optimal node to take at stage $i+1$ is
\begin{equation*}
    r^{i+1,\star}=\mathop{\arg\min}_{r^{i+1}\in\ch(r^i)} L_i(r^i,e^i)+\mathbb{E}_{e^{i+1}}[V(r^{i+1},e^{i+1})|e^i],
\end{equation*}
which is exactly $\pi^\star(r^i,e^i)$. This in turn shows that 
\begin{equation*}
\begin{aligned}
    V(r^i,e^i) &= L_i(r^i,e^i)+\mathbb{E}_{e^{i+1}}[V(r^{i+1,\star},e^{i+1})|e^i]\\
    &=\min_{r^{i+1}\in \ch(r^i)}Q(r^{i+1},e^i).
\end{aligned}
\end{equation*}
This is true for all nodes in $\bfr$ and $\bfe$, then by induction, $\pi^\star(s^i,e^i)$ is the optimal policy for all $i\in\{0,1,...,N\}$.
\end{proof}

\subsection{Extension to ego-conditioned prediction.}
\begin{figure}
    \centering
    \includegraphics[width=1\columnwidth]{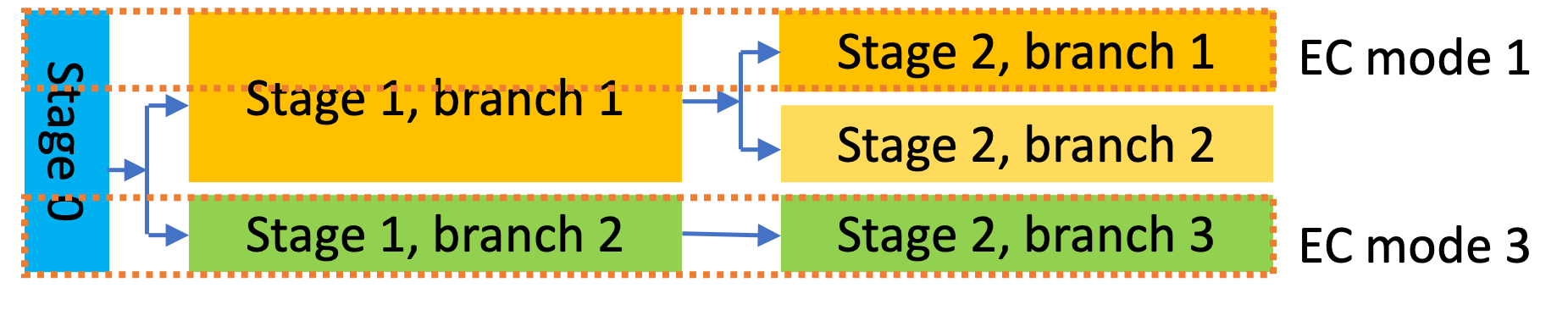}
    \caption{Flattened trajectory tree as ego-conditioning modes}
    \label{fig:EC}
    \vspace{-0.6cm}
\end{figure}

In the case where the scenario tree is ego-conditioned, the algorithm structure stays roughly the same. To generate the ego-conditioned trajectory prediction, we first run the ego motion sampler to generate the trajectory tree, then the trajectory tree is "flattened " to create the ego-conditioning (EC) modes. E.g., for trees in Fig. \ref{fig:diagram}, the flattened EC modes are shown in Fig. \ref{fig:EC}.
Under each EC mode, the prediction model generates a distinct scenario tree conditioned on the ego vehicle's motion. To keep the planning problem under ego-conditioning well-defined, the ensemble of EC scenario trees needs to satisfy a condition called causal consistency. 
\begin{definition}[Causal consistency]\label{def:consistency}
The ensemble of EC scenario tree is causally consistent (CC) if for any stage $i\le N$, any two scenario trees under different EC modes, if the conditioned ego trajectories are identical up to stage $i$, the two scenario trees are identical up to stage $i$.
\end{definition}

This condition guarantees the causality of the policy planning problem as for any particular stage $i$ within the prediction horizon, no future information beyond stage $i$ affects the prediction up to stage $i$. While causal consistency seems trivial, it may be violated by a multi-stage prediction model without special care. For example, if the EC trajectory is encoded as a whole for all stages, the scenario trees generated are not causally consistent.
\begin{prop}[Sufficient condition for CC]
Assume that a multi-stage ego-conditioned prediction model generates the trajectory prediction stage by stage and the EC prediction is deterministic, then the generated ensemble of EC scenario trees is causally consistent.
\end{prop}
\begin{remark}
For stochastic prediction models, the generated scenario trees are causally consistent if the random seeds are shared among all EC modes with the same EC trajectory.
\end{remark}

The proof is omitted here. It is straightforward to show that the 3 prediction models in Section \ref{sec:pred} are CC.

Given causally consistent EC predictions, only two terms change in the policy planning algorithm. First, all nodes of the scenario tree now depend on one of the nodes of trajectory tree of the same stage, and are denoted as $e^i(r^i)$ to emphasize the dependence. Note that the dependence is on $r^i$ instead of the whole ego trajectory due to causal consistency. Second, the branching probability now depends on the ego node, and is denoted as $\mathbb{P}[e^{i+1}_j|e^i(r^i),r^i]$.

Formally, we are solving a finite-horizon MDP with a joint state of $(r^i,e^i)$ and the cost defined in \eqref{eq:stage_cumu_cost}. The action space is $A(r^i)=\ch{r^i}$, the next ego maneuver, and the transition probability is 
\begin{equation*}
\resizebox{1\columnwidth}{!}{$
    \mathbb{P}[(r^{i+1},e^{i+1})|(r^i,e^i(r^i)),r^{i+1}_j] = 
    \left\{ \begin{aligned}
        &\mathbb{P}[e^{i+1}(r^{i+1}_j)|e^i(r^i),r^i],& \textbf{if}\; r^{i+1}=r^{i+1}_j\\
        &0,& \textbf{otherwise}
    \end{aligned} \right.
    $}
\end{equation*}
Interestingly, the value function can be defined exactly the same way as
\begin{equation}\label{eq:EC_V}
    V(r^i,e^i(r^i))=\mathbb{E}_{e^{i:N}} [\sum_{k=i}^N L_k(r^k,e^k(r^k))].
\end{equation}
Similarly, the Q function is now
\begin{equation*}
\resizebox{1\columnwidth}{!}{$
\begin{aligned}
    Q(r^{i+1},e^i(r^i))&:=L_i(r^i,e^i(r^i))+\mathbb{E}_{e^{i+1}}[V(r^{i+1},e^{i+1}(r^{i+1}))|e^i(r^i)]\\ &= L_i(r^i,e^i(r^i))+\sum_{e^{i+1}_j\in\ch(e^i(r^{i}),r^{i+1})} \mathbb{P}[e^{i+1}_j|e^i(r^i),r^i] V(r^{i+1},e^{i+1}_j),
    \end{aligned}
$}
\end{equation*}
where $\ch(e^i(r^{i}),r^{i+1})$ denotes the set of children nodes following $e^i(r^{i})$ on the tree associated with $r^{i+1}$.
The optimal policy is then defined as
\begin{equation}\label{eq:EC_opt_policy}
    \pi^\star(r^i,e^i(r^i))=\mathop{\arg\min}_{r^{i+1}\in\ch(r^i)} Q(r^{i+1},e^i(r^i)).
\end{equation}

\noindent\begin{thm}
Given a causally consistent ensemble of EC scenario trees, the policy in \eqref{eq:EC_opt_policy} is the optimal policy with respect to the cost function $\mathcal{J} = \sum_{i=0}^{N} L_i(r^i,e^i(r^i)).$
\end{thm}

\noindent\begin{proof}
The proof essentially follows the same steps as the proof of Theorem \ref{thm:DP}. Starting from stage $N$, the cost is constant, thus $\pi^\star$ is trivially optimal and $V(r^N,e^N(r^N))$ is the cost-to-go. For all $0\le i <N$, assume for all $r^{i+1}$, $V(r^{i+1},e^{i+1}(r^{i+1}))$ is the expected cost-to-go function at stage $i+1$. Then for any $r^i$, the cost-to-go to take any children node $r^{i+1}\in\ch(r^i)$ is $L_i(r^i,e^i(r^i))+\mathbb{E}[V(r^{i+1},e^{i+1}(r^{i+1}))|e^i(r^i)]$,
which is exactly $Q(r^{i+1},e^i(r^i))$. It follows that $\pi^\star(r^i,e^i(r^i))$ gives the optimal $b^{i+1}$ within $\ch(r^i)$ and the cost-to-go function of stage $i$ is computed as
\begin{equation*}
    V(r^i,e^i(r^i)) = \min_{r^{i+1}\in \ch(r^i)}Q(r^{i+1},e^i(r^i)).
\end{equation*}
The optimality of $\pi^\star$ is then proved by induction.
\end{proof}

\section{Experiments}\label{sec:experiment}

\begin{table}[]
\begin{tabular}{c|ccc}
                 & Rasterized & PredictionNet & Agentformer \\ \hline
Mean ADE      & 0.67       & 0.25          & 0.33        \\
Mean FDE      & 1.3        & 0.85          & 1.01        \\
\end{tabular}
\caption{Prediction metrics average displacement error (ADE) and final displacement error (FDE) of the three models. The mean is taken over all modes.}
\label{tab:prediction}
\vspace{-0.5cm}
\end{table}

\subsection{Training of prediction model}
We configure \algname{} with three different prediction models. Apart from the model architecture, the experimental setup is the same. Prediction models are trained with the nuScenes dataset \cite{caesar2020nuscenes}, comprised of 1000 scenes lasting for 20 seconds collected in urban areas of Boston and Singapore. For simplicity, only vehicles are considered. All prediction models use 2 stages and the branching factor is 4. 

\textbf{Loss function.} Our training loss is a weighted sum of the following terms: prediction loss that penalizes the error between the predicted trajectories and the ground truth; EC collision loss that penalizes collisions between the predicted trajectories and the conditioned ego trajectory;  collision loss that penalizes collisions between predicted trajectories of different agents in the scene; and additional standard regularization losses specific to the model architecture. 


\myparagraph{Ego conditioning.} One natural choice of the ego trajectory for ego-conditioning is the ground truth future trajectory. In addition, to expose the model to more diverse ego trajectories, we perturb the ground truth ego trajectory with noise generated via the \textit{Ornstein–Uhlenbeck} process as alternative ego trajectories. However, since the ground truth for surrounding agents under the counterfactual ego motion is not available, for the lack of a better choice, we still use the original ground truth as the target for the EC trajectory prediction. We penalize collision between the predicted trajectories of surrounding agents and the conditioned ego trajectory so that the model learns to avoid the conditioned ego trajectory.

\myparagraph{Training results.} We report standard ADE/FDE prediction metrics for the three prediction models in Table \ref{tab:prediction}. The results show that all three models learn reasonable predictions.

\begin{figure}
    \centering
    \includegraphics[width=0.8\columnwidth]{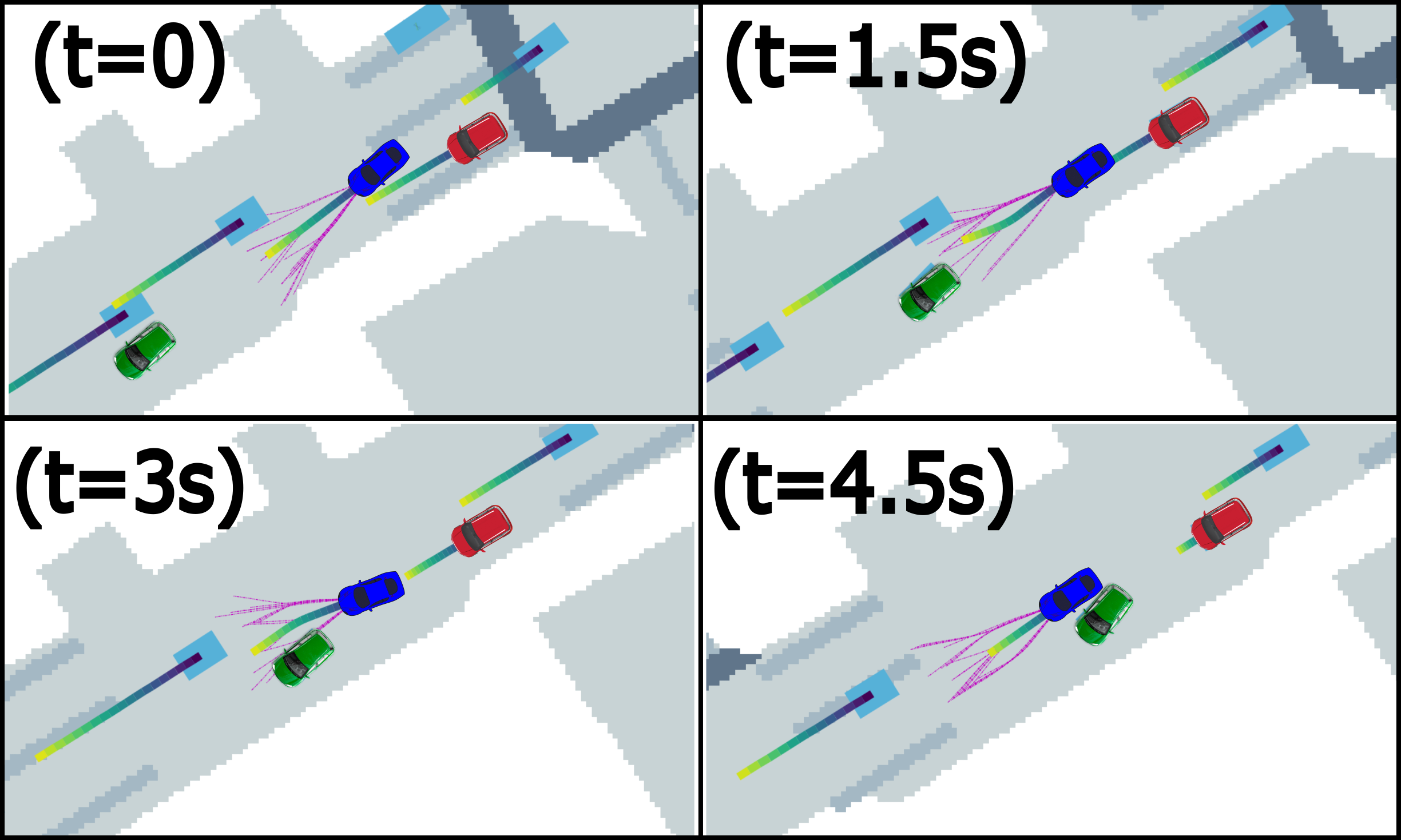}
    \caption{Example scene from simulation: the ego vehicle (blue) cuts in front of the red vehicle and then swerve around the green static vehicle in front. The ego trajectory tree is shown in magenta, the colored line shows agents' actual trajectory.}
    \label{fig:sim_snapshot}
    \vspace{-0.5cm}
\end{figure}

\subsection{Experimental setup}\label{sec:setup}

\myparagraph{Closed-loop simulation.} We evaluate \algname{} in BITS, a closed-loop AV simulation environment proposed in \cite{BITS}. Each simulation scene is initialized with a scene from the nuScenes dataset and then evolves forward with a policy for each agent. One agent is chosen as the ego vehicle and controlled by \algname{} or its alternatives. All other agents runs the learned BITS policy that generates realistic and diverse behavior of road vehicles. Additionally, to incite more interactions between agents, we "spawn" new agents around the ego vehicle to challenge the planner throughout the simulation duration.

\myparagraph{Metrics.} We use three key metrics: collision rate, offroad rate, and area coverage. The first two are calculated as the percentage of time steps where the ego vehicle is in collision (offroad). The coverage metric is calculated with a kernel density estimation (KDE) procedure following~\cite{BITS}, and it captures the effectiveness of the ego plan in terms of liveness and distance travelled.  
The evaluation is done on 100 scenes from the evaluation split of nuScenes (not used for training).

\myparagraph{Baselines.} We consider three baselines, the first two are non-contingent counterpart of \algname{}. For fair comparison, the two baselines share the same ego-motion sampler and ego-conditioning prediction model as \algname{}, the first baseline, named non-contingency robust (NCR), considers all possible trajectories predicted by the prediction model (similar to \cite{chen2020reactive}); while the second baseline, named non-policy greedy (NCG), only avoids the most likely prediction mode (similar to \cite{xu2014motion}). The third baseline uses a kinematic prediction model with 2 modes: maintaining speed and braking, and share the same multi-stage tree structure.

\bgroup
\setlength{\tabcolsep}{2pt}
\begin{table}[]
\vspace{-0.3cm}
\centering
\begin{tabular}{l|c|ccc}
 Planning  & Prediction   & Crash         & Offroad      & Coverage $\uparrow$\\ 
 algorithm & model       & rate $\downarrow$  & rate $\downarrow$ &   \\ 
\hline
TPP (ours) & Rasterized  & \textbf{0.46\%}     & \textbf{0.64\%}       & 164      \\
Non-Cont. Robust & Rasterized  & 0.80\%     & 0.61\%       & 163      \\
Non-Cont. Greedy & Rasterized  & 1.23\%     & 1.00\%       & \textbf{168}      \\
\hline
TPP (ours) & PredictionNet  & \textbf{1.02\%}     & \textbf{0.10\%}       &\textbf{174}      \\
Non-Cont. Robust & PredictionNet  & 1.34\%     & 2.07\%       & 166      \\
Non-Cont. Greedy & PredictionNet  & 1.60\%     & 0.45\%       & 170      \\
\hline
TPP (ours) & Agentformer  & \textbf{0.97\%}     & \textbf{0.20\%}       & \textbf{174}      \\
Non-Cont. Robust & Agentformer  & 1.08\%     & 0.63\%       & 158      \\
Non-Cont. Greedy & Agentformer  & 2.52\%     & 0.98\%       & 163      \\
\hline
TPP (ours) & Agentformer (Gaussian)  & 3.92\%     & 5.53\%       & 150      \\
Non-Cont. Robust & Agentformer (Gaussian)  & 1.40\%     & 0.40\%       & 144      \\
Non-Cont. Greedy & Agentformer (Gaussian) & 1.04\%     & 0.25\%       & 149      \\
\hline
TPP (ours) & Kinematic model & 1.75\% & 0.51\%  &160\\
\end{tabular}
\caption{Closed loop simulation results with \algname{} and the baselines, the default Agentformer model uses a discrete latent space and Agentformer (Gaussian) is its variant with a Gaussian latent space. \algname{} significantly outperform the baselines with the same ego trajectory tree and prediction model (bold means best).}
\label{table:metrics}
\vspace{-0.5cm}
\end{table}
\egroup

\subsection{Results}\label{sec:result}
Fig. \ref{fig:sim_snapshot}  shows snapshots of the simulation with \algname{} where the ego vehicle (blue) cuts in front of the red vehicle and then change lane again to avoid the parked green vehicle. The magenta lines show the ego trajectory tree, and the colorful line shows the rollout trajectories of the agents. The quantitative results are shown in Tables \ref{table:metrics} and \ref{table:EC}, and we make the following observations. Simulation videos can be found \href{https://sites.google.com/nvidia.com/avg-tpp/home}{here}.

\myparagraph{TPP outperforms the baselines.}
As shown in Table \ref{table:metrics}, \algname{} significantly outperforms the two non-policy baselines on crash and offroad rate under all 3 prediction models while achieving similar coverage. Even under a simple kinematic model, the performance is better than non-policy planners with sophisticated prediction models. \algname{} with deep-learned prediction models performs better than with a simple kinematic model, indicating the deep-learned prediction model benefits the closed-loop performance.

\myparagraph{Better predictions do not mean better plan.}
Interestingly, from Table \ref{table:metrics}, the "better" prediction model measured by ADE/FDE does not necessarily lead to better closed-loop performance. For example, the rasterized model is the simplest and "worst" model among the three, yet \algname{} with it outperforms the other two in terms of crash rate.

\myparagraph{Ego-conditioning is useful.} Table \ref{table:EC} compares \algname{} with or without ego-conditioning. Ego-conditioning improves the closed-loop performance considerably in most cases, especially the coverage metric, hinting that the ego vehicle is more comfortable moving through traffic when it expects reaction from other agents.

\myparagraph{Temporal consistency is important.} Additionally, we tested the Agentformer model with Gaussian latent space, where the prediction requires sampling of the latent variable at every inference call, compared to using discrete latent space where sampling is not needed. While the prediction ADE/FDE is similar, the closed-loop performance with the Gaussian latent is significantly worse. 
We suspect this is due to the poor temporal consistency between predictions of consecutive time steps. While both \algname{} and the two baselines saw performance degradation under the Gaussian latent, the policy planner relies on more sophisticated prediction/planning interaction, thus suffers more than the two baselines.

\begin{table}[]
\centering
\begin{tabular}{c|ccc}
                   & Crash rate$\downarrow$ & Offroad rate$\downarrow$ & Coverage $\uparrow$\\ \hline
Rasterized   & \textbf{0.46\%} / 0.53\%     & \textbf{0.60\%} / 1.05\%       & \textbf{164} / 163     \\
PredictionNet   & 1.00\% / \textbf{0.63\% }   & \textbf{0.10\%} / 0.71\%     & \textbf{174} / 151     \\
Agentformer  & \textbf{0.97\%} / 1.30\%    & \textbf{0.20\%} / 0.53\%      & \textbf{174} / 163     \\
\end{tabular}
\caption{\comment{this can be easily merged with Table III to save space } Ablation study of ego-conditioning, the first and second number are the metrics with and without ego conditioning. Most metrics see degradation when ego-conditioning is disabled (bold means better).}
\label{table:EC}
\vspace{-0.4cm}
\end{table}

\begin{table}[]
\centering
\begin{tabular}{l|ccc}
                    & Rasterized & PredictionNet & Agentformer \\ \hline
Prediction w/ EC    & 150ms      &   5.00s            &    2.60s        \\
Prediction w/o EC   &  37ms       & 75ms          & 150ms       \\
Ego-motion sampling & & 30ms &              \\
Dynamic programming & & 11ms &                
\end{tabular}
\caption{Runtime of the three key components}
\label{table:runtime}
\vspace{-0.5cm}
\end{table}

\myparagraph{Runtime analysis.}
Table \ref{table:runtime} shows the runtime of the three components of \algname{} on an Nvidia 3090 GPU (most computation is done on GPU). It should be noted that all runtime is recorded on unoptimized Python code and there is significant space of improvement. For example, the long runtime for PredictionNet is mainly due to rasterization, and we know its runtime can be as low as 5ms with the proper engineering \cite{predictionnet}, e.g., efficient CUDA code for rasterization. 
 While the runtime for prediction models is not very informative and is not the focus of this paper, it is quite clear that ego motion sampling and dynamic programming can be done very efficiently. With proper optimization, we believe that \algname{} can be easily implemented in real time. We plan to release the code upon publication.

\section{Conclusion}
We present \algname{}, a policy planner capable of generating multistage motion policies that react to the environment. It is centered around an ego trajectory tree and a scenario tree that predicts the environment behavior, both containing multiple stages. Ego-conditioning is applied to leverage the reactive behavior of the environment under the ego vehicle's presence. The closed-loop simulation result shows that \algname{} significantly outperform two non-policy benchmarks and the runtime test suggest that such sophisticated policy planner can be run in real time. Our planned future works include (1) study more flexible topology of the trees and enable event-triggered branching (2) improve the efficiency of computation, especially under ego-conditioning (3) adaptive and personalized policy planning.

\section*{Acknowledgement}
We would like to thank Bilal Kartal, Siva Hari, and Edward Schmerling for the insightful discussion.
\appendix

\subsection{Prediction model implementation}
\noindent\textbf{Rasterized model.} The rasterized model uses rasterized images as input, which contain the road geometry and the surrounding vehicles drawn as boxes around each agent. The image is passed through a CNN (RESNET) backbone, then through a transformer for message passing among agents. Finally, a conditional variational auto-encoder (CVAE) is used to generate multimodal predictions utilizing the dynamic models of the agents.
To generate multistage predictions, we run the same model $N$ times where $N$ is the number of stages. After each stage, all nodes of the same stage is batched with updated features (via a RNN feature roller) for the next stage.

\noindent\textbf{PredictionNet model.} The second example follows \cite{predictionnet} where the whole scene is painted on a single image, which is then processed by a Unet backbone. The prediction for each agent is obtained by sampling the output image at the location of the agent. This model is naturally scene-centric and takes the interaction between agents into account. To generate multi-stage predictions, a re-rasterization procedure is applied. Specifically, from stage 2 onward, we use the same background image as stage 1, and update the agent patches with the new agent history obtained by concatenating the original agent history with predictions from early stages.

\noindent\textbf{Agentformer model.}
The third example is adopted from the Agentformer model in \cite{agentformer}. Since it uses vectorized features, extending it to a multistage model is straightforward as we only need to take the moving window approach to obtain the agent history for later stages and encode them before feeding them to the transformer. A local map patch is passed through a CNN encoder as part of the feature for each agent, which does not change for all stages. To enable ego-conditioning, we simply remove the mask on the ego vehicle future so that the transformer has access to the information.

\bibliographystyle{IEEEtran}
\bibliography{bibliography}

\renewcommand{\baselinestretch}{0.97}

\end{document}